\documentclass{article}

\usepackage{arxiv}

\usepackage[utf8]{inputenc} 
\usepackage[T1]{fontenc}    
\usepackage{hyperref}       
\usepackage{url}            
\usepackage{booktabs}       
\usepackage{amsfonts}       
\usepackage{nicefrac}       
\usepackage{microtype}      
\usepackage{lipsum}
\usepackage{svg}
\usepackage{graphicx}
\graphicspath{ {./images/} }
\usepackage{amsmath} 
\usepackage{pdflscape}
\usepackage{xcolor}
\usepackage[normalem]{ulem}
\usepackage{comment}
\usepackage{longtable}
\usepackage{tabularx}
\usepackage{caption} 
\usepackage{subcaption}
\usepackage{longtable}
\usepackage{booktabs}
\usepackage[utf8]{inputenc}
\usepackage{changepage}

\usepackage{array}
\usepackage{float}
\newlength{\extralength}
\newlength{\fulllength}
\newcolumntype{C}{>{\centering\arraybackslash}X}

\title{Comparative Analysis of Generative Models: Enhancing Image Synthesis with VAEs, GANs, and Stable Diffusion}
\author{Sanchayan Vivekananthan  \\[1ex]
\begin{minipage}[t]{0.90\textwidth}
\centering
\scriptsize Department of Computer Science, Huddersfield University, Queensgate, Huddersfield HD1 3DH, UK; \\
Correspondence: U2380760@unimail.hud.ac.uk;
\end{minipage}}

\begin{document}

\maketitle
\begin{abstract} This paper examines three major generative modelling frameworks: Variational Autoencoders (VAEs), Generative Adversarial Networks (GANs), and Stable Diffusion models.VAEs are effective at learning latent representations but frequently yield blurry results. GANs can generate realistic images but face issues such as mode collapse. Stable Diffusion models, while producing high-quality images with strong semantic coherence, are demanding in terms of computational resources. Additionally, the paper explores how incorporating Grounding DINO and Grounded SAM with Stable Diffusion improves image accuracy by utilising sophisticated segmentation and inpainting techniques. The analysis guides on selecting suitable models for various applications and highlights areas for further research.
\end{abstract}


\section{Introduction}
In recent years, the field of generative modelling has seen remarkable progress, driven by the creation of advanced neural network architectures that can learn and generate complex data distributions. These models have reshaped fields like image synthesis, text generation, and speech synthesis while also paving the way for new research opportunities and practical applications in diverse areas, such as computer vision, natural language processing, and biomedical imaging ~\cite{aydin2023domain}.

Among the most prominent generative models are Variational Autoencoders (VAEs), Generative Adversarial Networks (GANs), and Stable Diffusion models. Each of these architectures provides distinct methods for learning data representations and creating new content, making them valuable tools for a variety of tasks, including image generation and anomaly detection. With the use of a probabilistic framework, VAEs can efficiently encode input into a latent space, making it easier to generate a variety of outputs. Although they have drawbacks such as mode collapse and unstable training dynamics, GANs have redefined the standard for producing realistic, high-quality images due to their adversarial training process. Stable Diffusion models, a more recent development, employ iterative refinement techniques to produce high-resolution and semantically consistent images, tackling some of the limitations present in VAEs and GANs.

Despite the success of these models, each comes with inherent limitations that can hinder their effectiveness in certain applications. For instance, VAEs often struggle with producing sharp images, GANs face challenges with training stability and diversity, and Stable Diffusion models, while producing high-quality outputs, are computationally intensive and time-consuming. To overcome these challenges, researchers have explored the integration of additional techniques, such as Grounding DINO and Grounded SAM, with Stable Diffusion. These techniques enhance the model’s ability to perform precise segmentation, object detection, and context-aware inpainting, thus broadening the applicability and effectiveness of generative models in more complex and nuanced tasks.

This paper aims to provide a comprehensive comparison of these leading generative modelling architectures, highlighting their strengths, limitations, and potential applications. Additionally, the paper explores the impact of integrating advanced techniques like Grounding DINO and Grounded SAM with Stable Diffusion, assessing how these enhancements can address the limitations of traditional models. Through this analysis, the paper seeks to guide researchers and practitioners in selecting the most suitable generative modelling architecture for their specific needs, while also identifying areas for future research and development in this rapidly evolving field.

\section{Variantional Autoencoder (VAE)} 

Variational Autoencoders (VAEs) are a popular neural network architecture for unsupervised learning of complex distributions. They combine the strengths of autoencoders with probabilistic techniques, making them effective for generative modelling and allowing them to create new data samples that closely resemble the training data \cite{ref2}.

\begin{figure}[H]
\begin{adjustwidth}{-\extralength}{0cm}
\centering
\includegraphics[height=3.5cm]{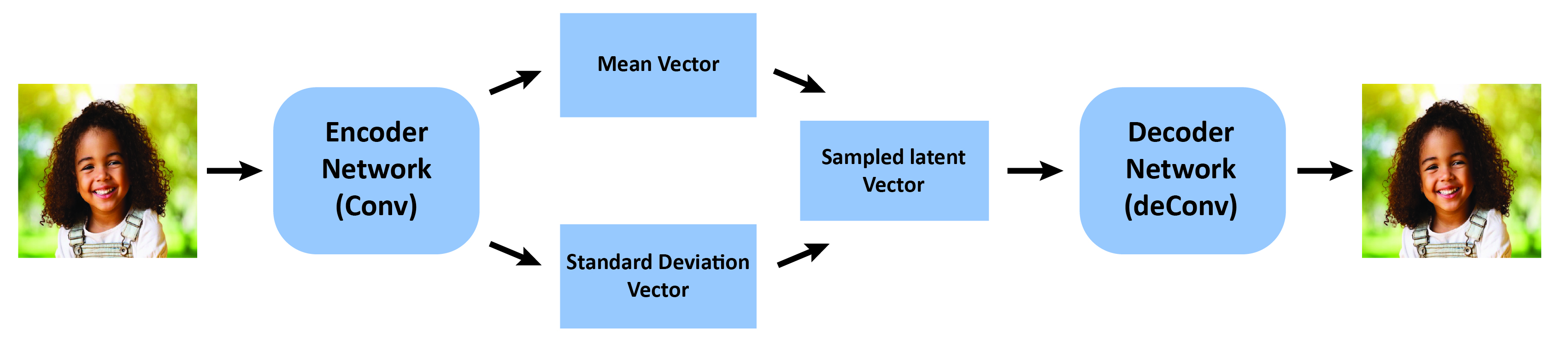}
\end{adjustwidth}
\caption{Variational Auto Encoder(VAE) architecture.}
\label{Figure:1}
\end{figure}

As shown in Figure \ref{Figure:1}, the architecture of a Variational Autoencoder (VAE) consists of two main components: an encoder and a decoder \cite{ref3}. The encoder transforms input data into a low-dimensional latent space, creating a "latent code" by producing key parameters like the mean and variance of a Gaussian distribution. The decoder then reconstructs the original data from this latent code. Depending on the data's nature, various neural network architectures, such as fully connected or convolutional networks, can be used for both the encoder and decoder. Through training, the VAE effectively captures essential data features, enabling accurate reconstruction. The reparameterization trick further optimizes the model using stochastic gradient descent \cite{ref4}.

Variational Autoencoders (VAEs) offer the advantage of learning intricate probability distributions from input data, making them highly effective for generative modelling tasks. By encoding data into a lower-dimensional latent space, VAEs facilitate efficient representation learning, which aids in understanding and disentangling the underlying factors of variation within the data. Additionally, they can be trained using gradient-based optimization techniques, which simplifies their implementation and allows them to scale well with large datasets. However, VAEs sometimes struggle with capturing fine details, leading to a fuzzy effect in reconstructions due to the use of L1 or L2 loss, which smooths out high-frequency details \cite{ref3}. Optimizing the prior distribution and achieving a balance between reconstruction accuracy and latent space regularization can be challenging \cite{ref5}. Moreover, VAEs are susceptible to posterior collapse, where the model disregards the latent variables, leading to less diverse and meaningless samples.

\section{Generative Adversarial Network (GAN)}
Generative Adversarial Networks (GANs) were first introduced by Ian Goodfellow and his colleagues in 2014 \cite{Gref1}. Since their introduction, GANs have been the subject of extensive research and have been praised for their innovative approach, especially in computer vision and image processing

\begin{figure}[H]
\begin{center}
\begin{adjustwidth}{-\extralength}{0cm}
\centering
\includegraphics[height=7.6cm]{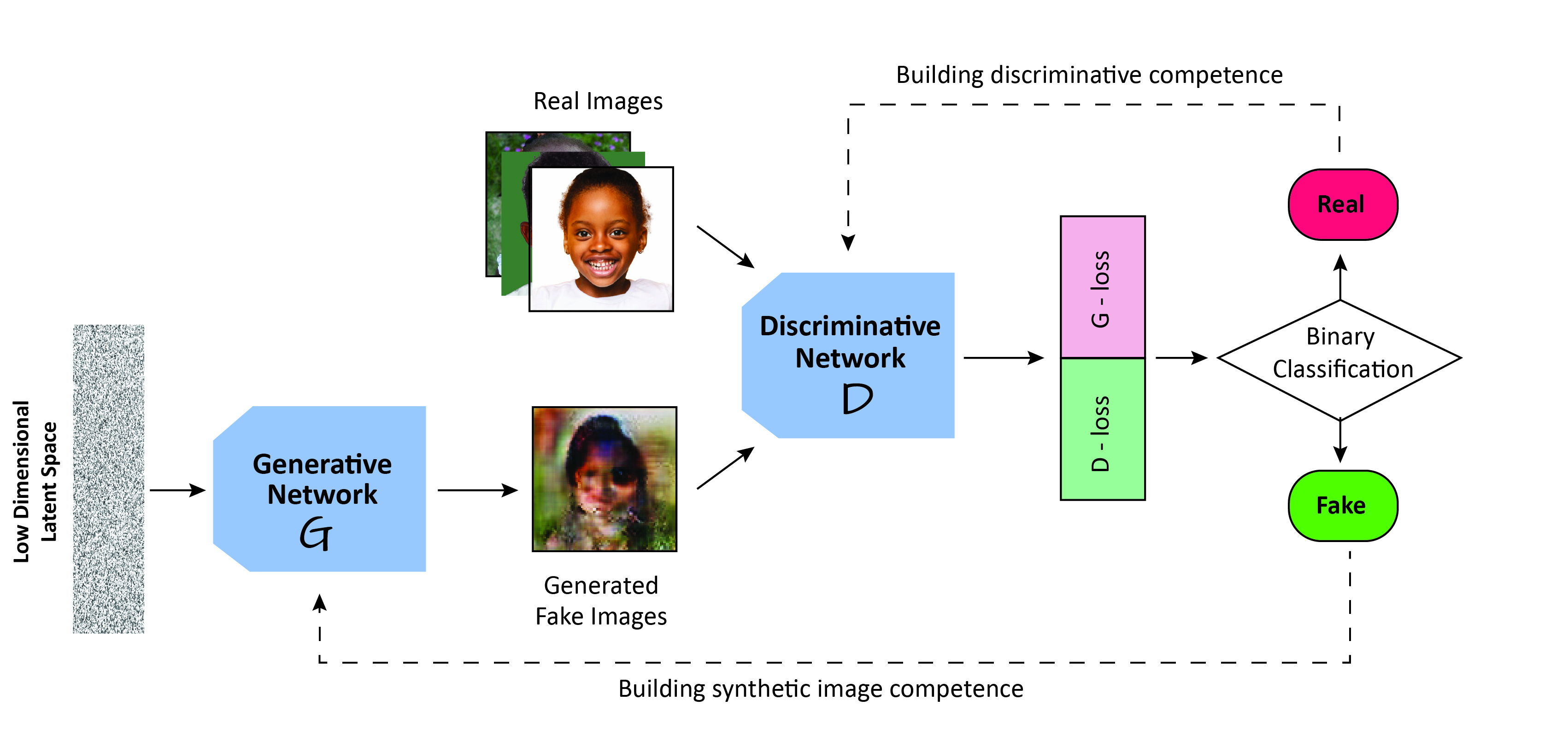}
\end{adjustwidth}
\caption{Generative Adversarial Network (GAN) Architecture}
\label{Figure:2}
\end{center}
\end{figure}

GANs are a generative model, meaning they can create new content based on the data they have been trained on. They have a wide range of applications, including high-resolution image quality enhancement \cite{Gref2}, artistic image editing \cite{Gref3}, realistic image generation \cite{Gref4}, cross-domain image transformation \cite{Gref5, hussain2023child}. However, their most popular application is in generating new images.

A Generative Adversarial Network (GAN) consists of two competing Neural Networks (NNs): the Generator and the Discriminator. As illustrated in Figure \ref{Figure:2}, the generator creates synthetic data that closely mimics real data, while the discriminator evaluates these outputs to determine whether they are genuine or fabricated. The generator's goal is to produce data that can convincingly pass as real, while the discriminator aims to accurately differentiate between real and fake data. This adversarial interaction helps the GAN improve iteratively, enhancing the realism and quality of the generated outputs over time.

There are several types of Generative Adversarial Networks (GANs), such as Vanilla, Conditional \cite{gref6}, Deep Convolutional \cite{gref7}, CycleGANs \cite{gref8}, StyleGANs \cite{gref9}, and Super Resolution GANs \cite{gref10}, each specifically designed for different applications in the field.

GANs, as part of the deep generative model (DGM) family, have rapidly gained popularity in the deep learning community due to several advantages over traditional DGMs. These advantages include:

\begin{itemize}
  \item Superior Output Quality: GANs have the capability to generate higher-quality outputs compared to other DGMs. Unlike the well-known variational autoencoders (VAEs), which struggle to produce sharp images, GANs can generate any form of probability density, resulting in more realistic outputs.
  \item Flexible Generator Training: The GAN framework allows for the training of any type of generator network, offering greater flexibility. In contrast, other DGMs often require specific conditions for the generator, such as having a Gaussian output layer, which can limit their applicability.
  \item Unrestricted Latent Variable Size: GANs impose no restrictions on the size of the latent variable, allowing for more adaptability in modelling complex data distributions.
\end{itemize}
These advantages have positioned GANs as a leading choice for generating synthetic data, particularly in image synthesis, where they consistently achieve state-of-the-art performance.

Even though GANs offer significant benefits in image synthesis, they come with certain limitations. Training can be unstable, often leading to issues such as mode collapse, where the model, in trying to fool the discriminator, repeatedly generates a few similar, unvaried outputs \cite{gref11}. Additionally, GANs require substantial computational resources, especially for high-resolution images, and may be overfitting training data, limiting the diversity of generated images. Evaluating their output is also subjective, with limited reliable metrics available.

\section{Stable diffusion}
While GANs struggle with mode collapse and adversarial issues, and VAEs often produce blurry images, Stable Diffusion overcomes these limitations by providing high-resolution, detailed, and diverse images, making it a superior alternative.

\begin{figure}[H]
\begin{adjustwidth}{-\extralength}{0cm}
\centering
\includegraphics[height=6cm]{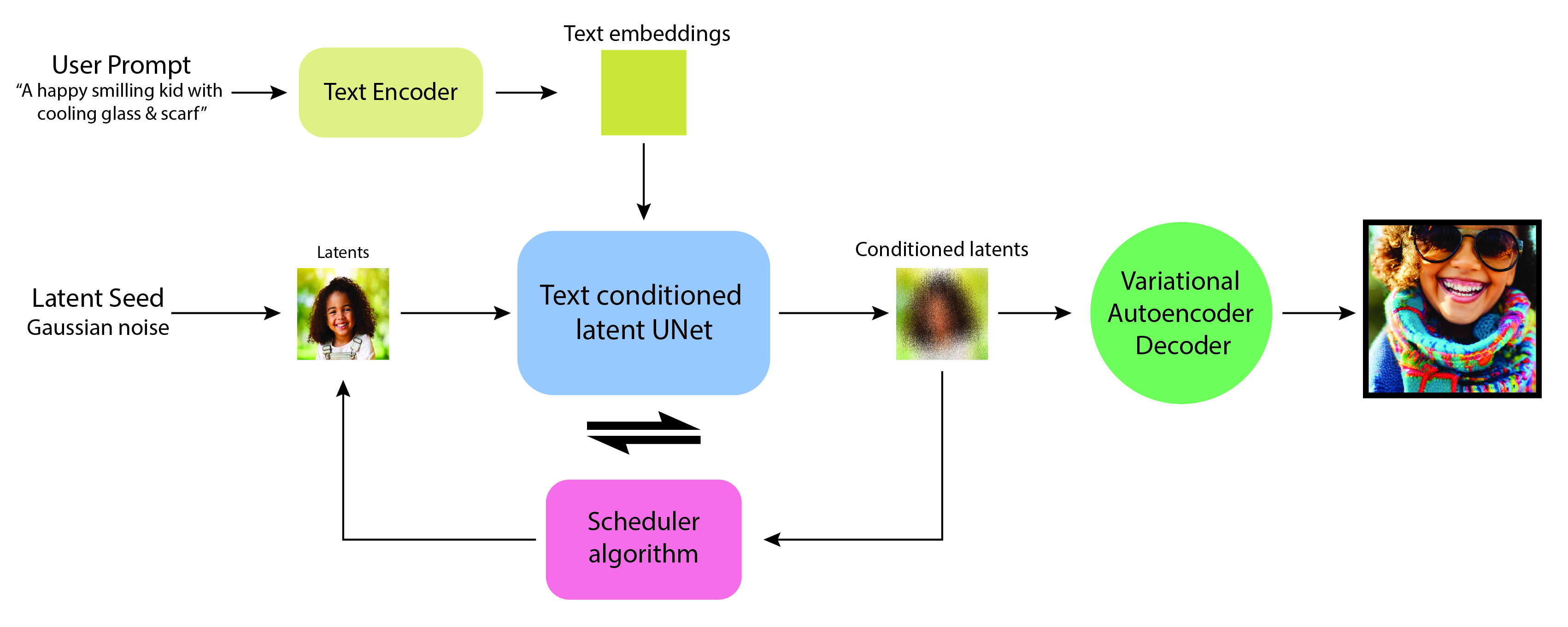}
\end{adjustwidth}
\caption{Stable Diffusion architecture.}
\label{Figure:3}
\end{figure}

Diffusion models have been utilized in numerous generative modelling tasks, including image generation \cite{sref1,sref2}, image-to-image translation \cite{sref3,sref4}, image editing\cite{sref5,sref6}, inpainting\cite{sref7,sref8}, and super-resolution \cite{sref9}, as shown in Figure \ref{Figure:3}. Additionally, the latent representations learned by these models have proven valuable for discriminative tasks such as image segmentation, classification, and anomaly detection.

These models synthesise images by reversing a progressive degradation process using three essential components: an autoencoder (VAE), a U-Net, and a text encoder. The process begins with the VAE, which compresses high-resolution images into lower-dimensional latent representations, significantly reducing computational demands. The training process consists of two stages: the forward diffusion phase and the reverse denoising phase \cite{sref10}. During the forward diffusion phase, noise is systematically introduced to the latent representations, progressively distorting them. In the subsequent reverse denoising phase, the U-Net predicts and removes this added noise to reconstruct the original image. The U-Net, with its encoder-decoder structure, operates on these noisy latents, utilizing multiple blocks, including ResNet layers and Vision Transformers, to refine and reconstruct the image. Additionally, a text encoder like CLIP converts input prompts into embeddings that guide the U-Net during the denoising process, ensuring that the generated images align with the textual descriptions\cite{sref11}. The final step involves the VAE decoder, which transforms the cleaned latent representation back into a high-resolution image.

Stable Diffusion provides several notable advantages for image generation and editing:
\begin{itemize}
  \item Superior Visual Quality: It produces images with fine details and lifelike textures through its iterative enhancement process.
  \item Consistent Semantic Integrity: The model maintains the original meaning of the input, resulting in images that are coherent and aligned with the intended content.
  \item Broad Feature Diversity: It effectively addresses the issue of mode collapse, offering a wide array of features and variations in the generated images, thus preventing repetitive results.
  \item Broad Applicability: It works well with both still and moving images and can handle different levels of noise, giving users greater flexibility in controlling the final output.
\end{itemize}

However, diffusion models face several limitations, with poor time efficiency during inference being a significant challenge. This inefficiency results from the requirement for numerous evaluation steps, often numbering in the thousands, to produce a single sample. Addressing this issue without sacrificing the quality of the generated samples is crucial for future progress. Several strategies can help overcome these challenges. One approach is to improve computational efficiency by developing more effective algorithms that reduce the computational load and speed up the sampling process. Optimizing sampling strategies is another key method, involving the exploration of new techniques that can maintain or even enhance sample quality while requiring fewer inference steps. Parallel processing techniques can also be leveraged to distribute the computational workload, accelerating the generation of samples. Additionally, incorporating prior knowledge or domain-specific information into the diffusion process can lead to more efficient outcomes. 

\section{Grounding DINO coupled with SAM and Stable diffusion inpainting}
In the evolving field of image synthesis, Stable Diffusion has proven to be a powerful tool for generating high-quality, detailed images. However, by integrating advanced techniques like Grounding DINO and Grounded SAM, the capabilities of Stable Diffusion can be significantly enhanced \cite{dref1}. These methods introduce precise segmentation, object detection, and contextual awareness into the synthesis process, enabling more accurate and contextually coherent image generation.

\begin{figure}[H]
\begin{center}
\begin{adjustwidth}{-\extralength}{0cm}
\centering
\includegraphics[height=7.6cm]{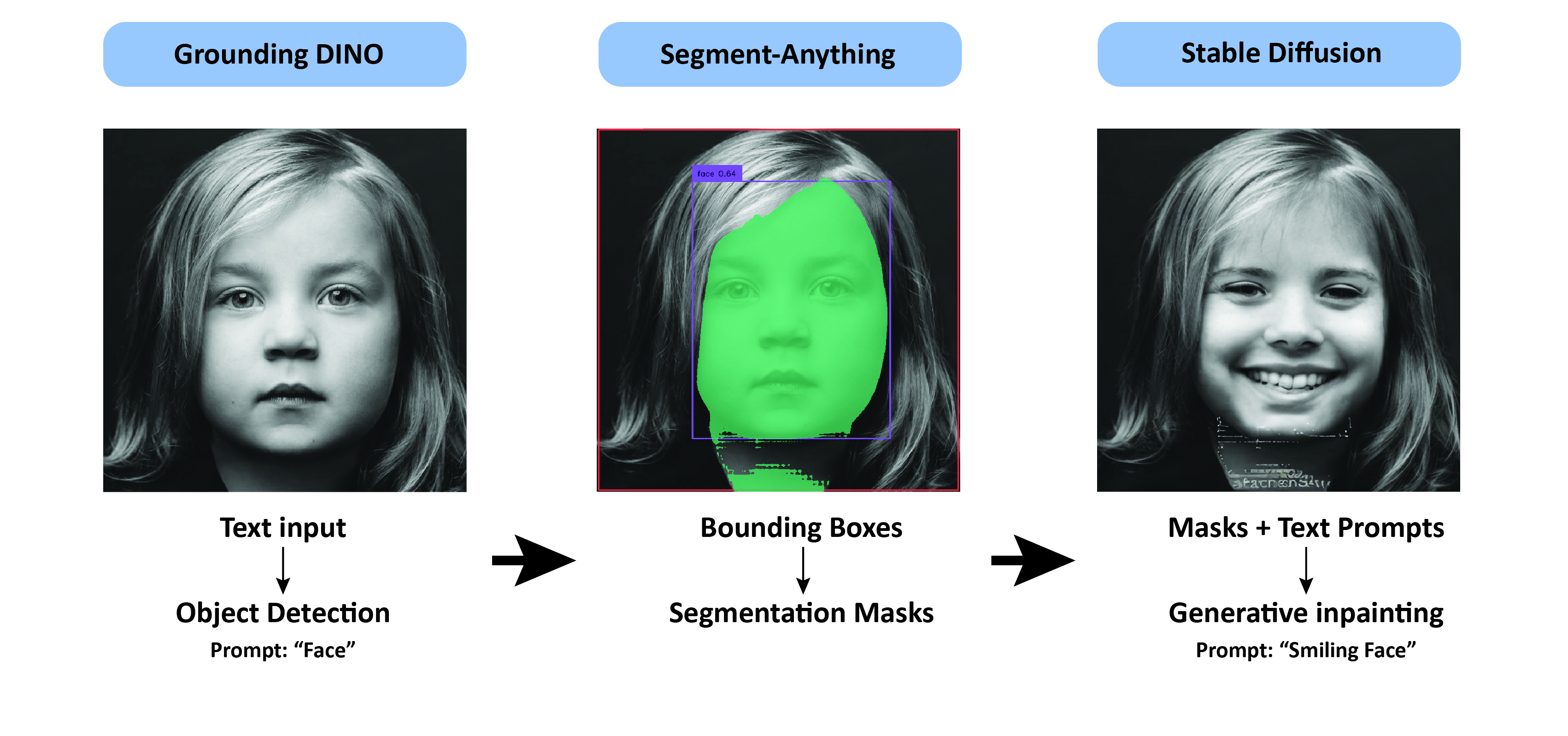}
\end{adjustwidth}
\caption{Advanced Image Synthesis with Grounded DINO, SAM and Stable Diffusion inpainting}
\label{Figure:4}
\end{center}
\end{figure}

\begin{itemize}
  \item Grounding DINO: Grounding DINO leverages advanced object detection and grounding techniques to further refine the masks and provide contextual information. It ensures that the inpainting process aligns with the content and structure of the image.
  \item Grounded SAM (Segment Anything Model): Grounded SAM is used to create precise segmentation masks based on specific regions of interest in the image. This allows for accurate localization and targeting of the areas that need inpainting.
  \item Stable Diffusion Inpainting: Stable Diffusion then uses these refined masks and contextual information to generate high-quality inpainted areas. By iteratively refining the noisy input, Stable Diffusion fills in missing parts with realistic details that seamlessly blend with the surrounding content.
\end{itemize}
Integrating Grounded SAM and Grounding DINO with Stable Diffusion, as shown in Figure \ref{Figure:4} offers significant advantages, including enhanced precision through accurate segmentation, improved contextual awareness, and higher-quality image outputs with better detail and coherence. However, these benefits come with trade-offs. The process becomes more complex, requiring greater computational resources and expertise, and may result in longer processing times. Additionally, the use of specialized techniques could increase the risk of overfitting, potentially limiting the method's applicability across diverse image types. A further challenge arises when trying to detect and modify very small or less visible areas in an image, as the prompt may struggle to accurately identify and target these subtle details.

\section{Discussion}

In this paper, we have explored and compared three major generative modelling architectures: Variational Autoencoders (VAEs), Generative Adversarial Networks (GANs), and Stable Diffusion, along with an advanced integration approach that combines Stable Diffusion with Grounding DINO and Grounded SAM for enhanced image synthesis. Each of these models has distinct advantages and limitations, which makes them suitable for different applications within the field of generative modelling.

Variational Autoencoders (VAEs) provide an efficient representation of the underlying structure in a lower-dimensional latent space and are useful for learning complex probability distributions from data. Their probabilistic framework enables the creation of varied outputs from a continuous latent space. However, VAEs often yield blurry reconstructions because L1 or L2 loss functions tend to smooth out high-frequency details. Additionally, problems like posterior collapse and the difficulty in balancing reconstruction accuracy with latent space regularization are significant challenges.

Generative Adversarial Networks (GANs), in contrast, have demonstrated impressive success in creating high-quality, lifelike images. Their adversarial training method enables them to produce sharp, detailed outputs that often exceed the quality of those generated by VAEs. However, GANs face challenges like mode collapse, where the generator produces a narrow range of repetitive outputs, and training instability, which can be costly in terms of computation. Additionally, the subjective nature of evaluating images generated by GANs makes it difficult to consistently assess their performance.

Stable Diffusion represents a more recent and advanced approach, offering high-resolution, detailed images with a robust handling of diverse features. By leveraging a progressive degradation and reconstruction process, Stable Diffusion can overcome the limitations of both VAEs and GANs, particularly in terms of maintaining visual fidelity and feature diversity. Nonetheless, diffusion models require a significant number of computational steps to generate each image, which can be time-consuming and resource-intensive.

The integration of Grounding DINO and Grounded SAM with Stable Diffusion further enhances the capabilities of image synthesis by introducing precise segmentation and contextual awareness. This makes it possible to generate images with more accuracy and coherence, particularly for tasks like object localisation and inpainting. However, the complexity, computational load, and risk of overfitting associated with this sophisticated integration may limit its applicability to varying types of images.

Table \ref{table:architectures} summarizes the key features, advantages, and limitations of the discussed architectures:

\begin{table}[H]
\centering
\caption{Comparison of Generative Modeling Architectures}
\label{table:architectures}
\begin{tabular}{|p{3.5cm}|p{3.5cm}|p{4cm}|p{4cm}|}
\hline
\textbf{Architecture} & \textbf{Key Features} & \textbf{Advantages} & \textbf{Limitations} \\ \hline
\textbf{Variational Autoencoders (VAEs)} & Latent space encoder-decoder structure; probabilistic model & Efficient representation learning; continuous latent space & Reconstructions with blurry edges; posterior collapse; balance between regularisation and accuracy\\ \hline
\textbf{Generative Adversarial Networks (GANs)} & Adversarial training between Generator and Discriminator & High-quality, realistic images; flexible generator training & Mode collapse; unstable training; subjective evaluation; computationally intensive \\ \hline
\textbf{Stable Diffusion} & Progressive degradation and reconstruction; autoencoder, U-Net, text encoder & feature diversity; consistent semantic integrity and comprehensive; high-resolution images & Time-consuming during inference; high computational cost; complex optimization \\ \hline
\textbf{Grounding DINO \& Grounded SAM with Stable Diffusion} & Object detection, segmentation, contextual inpainting & Increased accuracy; superior detail and coherence; enhanced contextual understanding & Greater complexity; increased demand for computational resources; potential for overfitting; extended processing times \\ \hline
\end{tabular}
\end{table}

\section{Conclusions}
In conclusion, while each of these generative models has its strengths, the choice of architecture depends on the specific application requirements. VAEs are well-suited for tasks requiring efficient latent space representation, GANs excel in high-quality image synthesis, and Stable Diffusion, especially when integrated with advanced techniques like Grounding DINO and Grounded SAM, which offer superior image generation capabilities. However, the trade-offs in computational resources, complexity, and generalizability must be carefully considered when selecting or developing a generative model for specific use cases such as ~\cite{animashaun2023automated,hussain2023exudate,zahid2023lightweight}. Future research should focus on overcoming the limitations of these models, such as improving the efficiency of diffusion models or stabilizing GAN training, to further advance the field of generative modelling.

\begin{adjustwidth}{-\extralength}{0cm}

\bibliographystyle{unsrt}  
\bibliography{my_references}  

\end{adjustwidth}
\end{document}